\pdfoutput=1

\documentclass[11pt]{article}

\usepackage{acl}

\usepackage{times}
\usepackage{latexsym}
\usepackage{algorithm}
\usepackage{algorithmic}
\usepackage{footnote}
\usepackage{url}
\usepackage{multirow}
\usepackage[switch]{lineno}
\usepackage{booktabs}
\usepackage{makecell}

\usepackage[T1]{fontenc}

\usepackage[utf8]{inputenc}

\usepackage{microtype}
\usepackage{graphicx}
\usepackage{amsmath}
\usepackage{bbm}

\newcommand{\fref}[1]{Figure~\ref{#1}}
\newcommand{\tref}[1]{Table~\ref{#1}}
\newcommand{\sref}[1]{\S\ref{#1}}

\title{SCAT: Robust Self-supervised Contrastive Learning via Adversarial Training for Text Classification}

 \author{Junjie Wu \quad Dit-Yan Yeung \\
         The Hong Kong University of Science and Technology}

\begin{document}
\maketitle
\begin{abstract}
Despite their promising performance across various natural language processing (NLP) tasks, current NLP systems are vulnerable to textual adversarial attacks. To defend against these attacks, most existing methods apply adversarial training by incorporating adversarial examples. However, these methods have to rely on ground-truth labels to generate adversarial examples, rendering it impractical for large-scale model pre-training which is commonly used nowadays for NLP and many other tasks. In this paper, we propose a novel learning framework called SCAT (\textbf{S}elf-supervised \textbf{C}ontrastive Learning via \textbf{A}dversarial \textbf{T}raining), which can learn robust representations without requiring labeled data. 
Specifically, SCAT modifies random augmentations of the data in a fully label-free manner to generate adversarial examples.
Adversarial training is achieved by minimizing the contrastive loss between the augmentations and their adversarial counterparts. We evaluate SCAT on two text classification datasets using two state-of-the-art attack schemes proposed recently. Our results show that SCAT can not only train robust language models from scratch, but it can also significantly improve the robustness of existing pre-trained language models. Moreover, to demonstrate its flexibility, we show that SCAT can also be combined with supervised adversarial training to further enhance model robustness.
\end{abstract}

\section{Introduction}

Deep learning models, especially Pre-trained Language Models \cite{devlin2019bert,lan2019albert,raffel2020exploring}, have achieved great success in the NLP field. However, a critical problem revealed by recent work is that these models are vulnerable to small adversarial perturbations \cite{kurakin2016adversarial,alzantot2018generating,ren2019generating,jin2020bert,li2020bert}. Attackers can simply fool elaborately fine-tuned models by substituting a few tokens in the input text or adding perturbations in the embedding space. Hence, there exist urgent needs for enhancing the robustness of current language models.

Existing defense schemes in the text domain can be categorized into three main groups. The first one is adversarial training \cite{miyato2016adversarial,zhu2020freelb,wang2020infobert}, which has been widely used in computer vision (CV) tasks \cite{goodfellow2014explaining,madry2018towards}. However, these methods typically fail to guard textual attacks due to the discrete nature of text data. To solve this problem, the second group of methods apply discrete adversarial examples instead to perform adversarial training \cite{jia2017adversarial,wang2018robust,liu2020robust, bao2021defending}. Lastly, some carefully designed models are proven to be effective when defending against specific attacks \cite{jia2019certified,jones2020robust,le2021sweet}. 
However, most of these methods need ground-truth labels when generating adversarial examples, making them unsuitable for large-scale model pre-training which is a popular scheme these days. 
Also, specially designed models like those proposed by \citet{jones2020robust} could not generalize well to state-of-the-art attacks \cite{jin2020bert,li2020bert}.

Recently, 
contrastive learning has become a popular paradigm to obtain high-quality representations for different downstream tasks \cite{chen2020simple,giorgi2020declutr}. Motivated 
by its great success, some CV 
researchers proposed ``label-free” adversarial pre-training methods to enhance model robustness \cite{kim2020adversarial, jiang2020robust}. However, these 
frameworks rely on gradient-based attacks on continuous pixels and thus cannot be transferred in a straightforward way to NLP tasks.

To tackle the above problems, in this paper, we propose a novel learning framework named SCAT (\textbf{S}elf-supervised \textbf{C}ontrastive Learning via \textbf{A}dversarial \textbf{T}raining). More concretely, we first apply a naive yet useful data augmentation method \cite{lowell2020unsupervised} to obtain different views of the original input. We then create adversarial examples \footnote{We term adversarial examples used in SCAT pre-training as ``label-free adversarial examples'' in the rest of the paper, in order to distinguish them from those generated by the two attackers we use in experiments.}
by replacing tokens in the augmentations via a masked language model. Adversarial training 
is performed by minimizing the contrastive loss between the augmentations and their adversarial counterparts. It is noteworthy that no ground-truth labels are needed in the above process. 

We evaluate the effectiveness of SCAT on two different text classification datasets using two state-of-the-art textual attackers. Our results show that SCAT can not only pre-train robust models from scratch, but it can also significantly enhance the robustness of existing pre-trained language models such as BERT \cite{devlin2019bert}. Further performance gain in terms of robustness obtained when combining SCAT with supervised adversarial training also demonstrates the flexibility of SCAT.

The contributions of this paper are as follows: 
\begin{itemize}
\setlength{\itemsep}{0.5pt}
\setlength{\parsep}{0.5pt}
\setlength{\parskip}{0.5pt}
    \item We propose an efficient token substitution-based strategy to create adversarial examples for textual inputs in a fully label-free manner.
    \item To the best of our knowledge, our proposed \textbf{SCAT} framework is the first attempt to perform textual adversarial training via contrastive learning without using labeled data.  
    \item Extensive experimental results demonstrate that SCAT can endow different models a significant boost of robustness against strong attacks in flexible ways. 
\end{itemize}

\section{Related Work}

\textbf{Textual Adversarial Attacks.}\quad Compared to the success of adversarial attacks in CV, attacking textual data is more challenging due to its discrete nature. Some early works \cite{papernot2016crafting,zhao2018generating} adapt gradient-based attacks to text data by introducing perturbations in the high-dimensional embedding spaces directly. However, these attacks lack semantic consistency since there do not exist obvious relationships between words and their embedding values. To make the generated adversarial examples fluent and similar to their original counterparts, recent NLP attacks focus on designing heuristic rules to edit characters \cite{belinkov2018synthetic,eger2019text}, words \cite{alzantot2018generating,ren2019generating,jin2020bert,li2020bert, yoo2020searching,li2020contextualized} or sentence segments \cite{jia2017adversarial,han2020adversarial,xu2021grey,lin2021using} in the input sequence and achieve great performance on different benchmarks.

\noindent \textbf{Defense Schemes in NLP.}\quad Various defense methods have been designed for NLP systems. 
One of them is adversarial training, which has been a prevailing way to improve model robustness in CV \cite{goodfellow2014explaining,madry2018towards}. To fill the gap between discrete text and continuous pixels when transferring this approach to NLP, researchers either attempt to add continuous perturbations to high-dimensional spaces like word embedding \cite{miyato2016adversarial,sato2018interpretable,zhu2020freelb,li2020textat,wang2020infobert,si2021better, guo2021gradient}, or generate discrete adversarial examples first and then perform adversarial training \cite{jia2017adversarial,wang2018robust, ivgi2021achieving, bao2021defending}. Several elaborately designed models
have also been proposed to defend specific attacks such as character-level typos \cite{pruthi2019combating,zhou2019learning,jones2020robust,ma2020charbert} and token-level substitutions \cite{zhou2019learning,wang2020infobert,si2021better}. More recently, certified-robustness \cite{jia2019certified,wang-etal-2021-certified} has attracted much attention since it ensures the failure of adversarial attacks to some extent via the interval bound propagation method. Nevertheless, these methods still have much room for improvement. As pointed out by \citet{li2020textat}, gradient-based adversarial training is not suitable for textual data. Also, most of these defense methods need to touch ground-truth labels, but label scarcity is a common problem in many machine learning tasks due to the costly human annotation procedure. Hence, it is crucial to design label-free defenses for NLP systems.

\noindent\textbf{Contrastive Learning and Adversarial Robustness.}\quad Self-supervised learning, especially contrastive learning, is becoming popular since it can generate high-quality representations over different modalities without using class labels \cite{devlin2019bert,chen2020simple,grill2020bootstrap,fang2020cert,wu2020clear,giorgi2020declutr,gao2021simcse,liu2021dialoguecse}. Some recent studies demonstrated that unlabeled data could be used to generate robust representations for images \cite{carmon2019unlabeled,hendrycks2019using}. Later on, following the SimCLR framework \cite{chen2020simple}, \citet{kim2020adversarial} and \citet{jiang2020robust} proposed label-free approaches to craft gradient-based adversarial examples for adversarial contrastive learning. However, to the best of our knowledge, no efforts in the NLP fields have been made in this direction, mainly due to the subtle differences between vision and language. Contrariwise, our proposed SCAT framework solves the above issues in a general and flexible paradigm 
and can defend various textual attacks effectively.

\section{Methodology}
\label{method}
\subsection{Preliminaries}
\label{pre}
\noindent\textbf{Self-supervised Contrastive Learning.} \quad Introduced by \citet{chen2020simple}, SimCLR can achieve similar performance compared to models trained for image classification tasks in a supervised manner. The workflow of SimCLR can be briefly explained as follows: considering a randomly sampled minibatch $\{x_{1}, x_{2}, \ldots, x_{N}\}$. For each example $x_{i}$, a data augmentation module is applied to yield two correlated views denoted by $x_{i}^{(1)}$ and $x_{i}^{(2)}$. Let $B$ denote the set of 2$N$ augmented examples thus formed from the original minibatch. For each augmented example $x_{i}^{(1)} \in B$, it forms a positive pair with $x_{i}^{(2)}$ and $2(N-1)$ negative pairs with the other augmented examples in the set $B\backslash \{x_{i}^{(1)}, x_{i}^{(2)}\}$. Let the encoder and projection head be $f(\cdot)$ and $g(\cdot)$, respectively. They map $x_{i}^{(1)}$ and $x_{i}^{(2)}$ to hidden representations $z_{i}^{(1)}$ and $z_{i}^{(2)}$ as $z=g(f(x))$. We can thus define the contrastive loss function for a positive pair $(x_{i}^{(1)}, x_{i}^{(2)})$ as: 

\begin{small}
\begin{linenomath}
\begin{align}
\label{eq1}
    l_{(x_{i}^{(1)}, x_{i}^{(2)})} &= {\rm -log} \frac{{\rm exp}({\rm sim}(z_{i}^{(1)}, z_{i}^{(2)})/\tau)}{\sum_{x_{j} \in B\backslash \{x_{i}^{(1)}\}}{
    \rm exp}({\rm sim}(z_{i}^{(1)}, z_{j})/\tau)}\\
    \label{eq2}
    L_{\text{CL}_{(x_{i}^{(1)}, x_{i}^{(2)})}} &= l_{(x_{i}^{(1)}, x_{i}^{(2)})} + l_{(x_{i}^{(2)}, x_{i}^{(1)})}
\end{align}
\end{linenomath}
\end{small}

\noindent where 
$z_{j} = g(f(x_{j}))$. ${\rm sim}(x,y)$ denotes the cosine similarity between two vectors, and $\tau$ is a temperature parameter. We define $\lbrace x_{i_{pos}}\rbrace$ as the positive set including the augmentations of $x_{i}$ and $\lbrace x_{i_{neg}} \rbrace$ the negative set containing the augmentations of other instances. 

\noindent\textbf{Adversarial Training.} \quad Adversarial training is a prevalent method for enhancing model robustness. The key idea behind it is solving a min-max optimization problem. Details about adversarial training's key idea can be found in Appendix~\ref{appendix:adversarial training}. Overall, our model is built upon 
these preliminaries. More details will be described below.


\subsection{Problem Definition and Model Overview}
Formally, given a set of data $\{(x_{i},y_{i})\}_{i=1}^{N}$ and a text classification model $M:X\to Y$, 
adversarial attackers aim to make $M$ give wrong predictions with slight modifications on $x_{i}$. In this paper, we want to build a robust model $M_{R}$ that can still generate correct labels against such perturbations in a fully label-free manner. 

\begin{figure}
    \centering
    \includegraphics[width=0.48\textwidth]{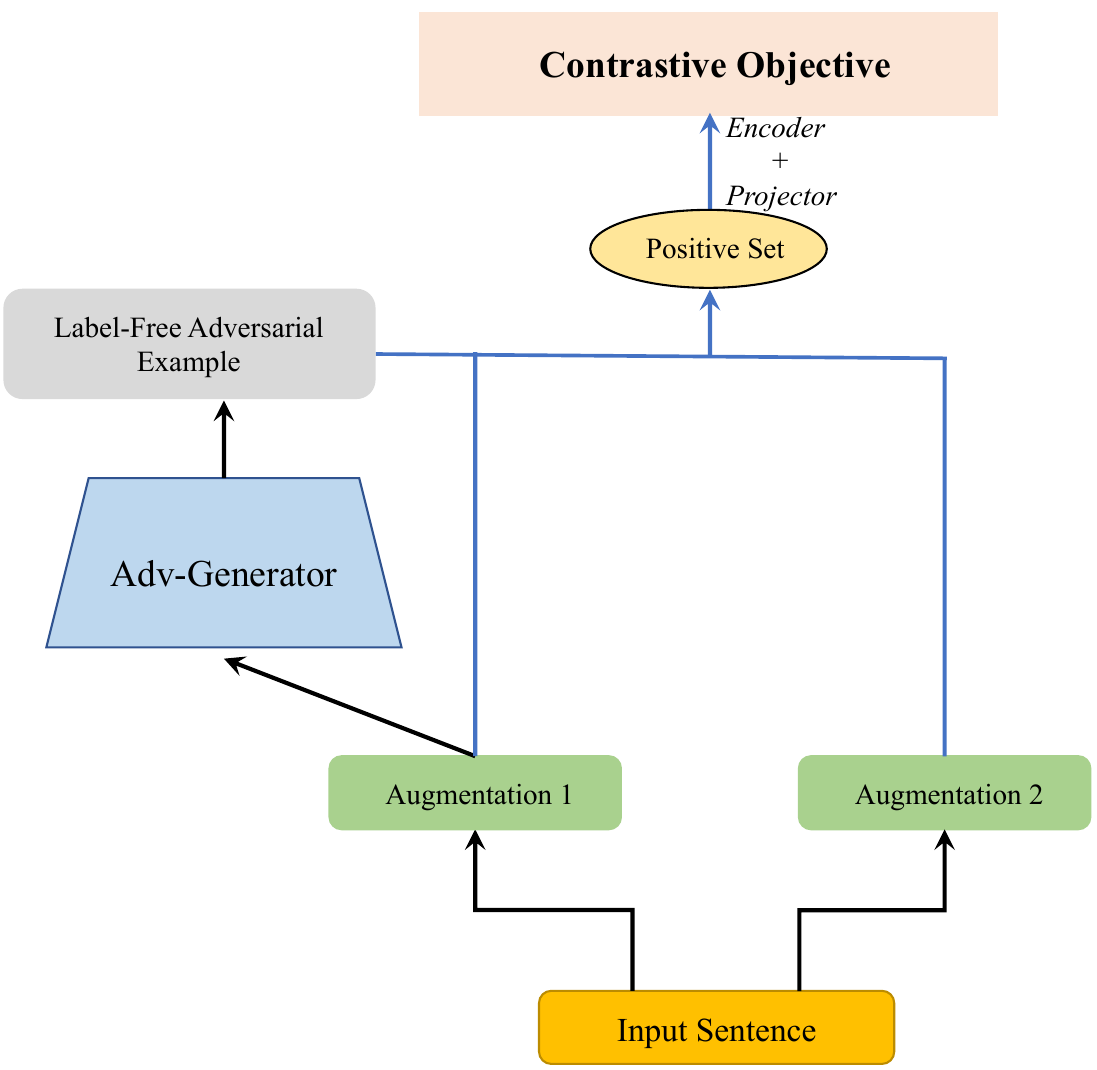}
    \caption{Overview of our proposed SCAT framework. The top blue arrow represents the encoding then projecting process for a given sequence.}
    \label{fig:model overview}
\end{figure}
\fref{fig:model overview} illustrates the workflow of our proposed method.
Given an input sentence, SCAT first randomly generates two augmented examples via a data augmentation module (\sref{data augmentation}). Then, the Adv-Generator will craft a label-free adversarial example by substituting a few tokens in one of the augmentations. This step utilizes the gradient information provided by the contrastive loss between the two augmentations. Next, we add this label-free adversarial example to the original positive set as a regularizer to help our model defend malevolent attacks. Finally, we perform adversarial training 
by minimizing the contrastive loss between the two augmentations and the adversarial counterpart.

\begin{savenotes}
\begin{algorithm*}
	\renewcommand{\algorithmicrequire}{\textbf{Input:}}
	\renewcommand{\algorithmicensure}{\textbf{Output:}}
	\caption{Label-free Adversarial Example Generation}
	\label{alg:1}
	  \algsetup{linenosize=\small}
    \small
	\begin{algorithmic}[1]
		\REQUIRE Two augmentations of input $x_{i}$: $T(x_{i})_{1}$, $T(x_{i})_{2}$, encoder $E$, projector $P$, attack percentage $\epsilon$, constant
		$K$, masked language model $\mathit{MLM}$.
		\ENSURE Label-free Adversarial example $x_{i}^{adv}$, contrastive loss $L_{\text{CL}(x_{i}^{adv}, T(x_{i})_{2})}$. 
		\STATE Copy $T(x_{i})_{1}\Longrightarrow T(x_{i})^{c}_{1}$  
		\STATE 
		$T(x_{i})^{c}_{1} = (w_{1}, w_{2}, ..., w_{n})$
		\STATE Calculate 
		$L_{CL}$ between two augmentations using Eq.\ref{eq1} and Eq.\ref{eq2}
		\FOR{$w_{j}$ in $T(x_{i})^{c}_{1}$}
		    \STATE 
		    Get its importance score $I_{w_{j}}$ using 
		    gradient information
		\ENDFOR
	    \STATE Sort tokens in $T(x_{i})^{c}_{1}$ in descending order with $I_{w_{j}}$ 
	    \STATE Pick top-$\epsilon$ words from the sorted sequence $\Longrightarrow$ $\mathit{ATT} = \lbrace w_{1}, w_{2}, ..., w_{k}\rbrace$
	    \STATE $\mathit{ATT}_{sub} = \lbrace s_{1}, s_{2}, ..., s_{N}\rbrace$ \COMMENT{sub-word tokenized set of $\mathit{ATT}$}
	    \STATE Use $\mathit{MLM}$ to generate top-K substitutions for tokens in $\mathit{ATT}_{sub}$
	    \FOR{$w_{p}$ in $\mathit{ATT}$}
	        \IF{$w_{p}$ is an intact word in $\mathit{ATT}_{sub}$}
	            \STATE 
	            Get candidates $C$ filtered by semantic constraints
	        \ELSE
	            \STATE 
	            Get $C$ filtered by PPL and semantic constraints
	        \ENDIF
	        \STATE Randomly pick one 
	        item $c_{q}$ from $C$
	        \STATE Replace $w_{p}$ with $c_{q}$ in $T(x_{i})^{c}_{1}$
	    \ENDFOR
	    \STATE $x_{i}^{adv} = T(x_{i})^{c}_{1}$
	    \STATE Calculate $L_{\text{CL}(x_{i}^{adv}, T(x_{i})_{2})}$ using Eq.\ref{eq1} and Eq.\ref{eq2}
		\STATE \textbf{return} $x_{i}^{adv}$, $L_{\text{CL}(x_{i}^{adv}, T(x_{i})_{2})}$
	\end{algorithmic}  
\end{algorithm*}
\end{savenotes}

\subsection{Data Augmentation}
\label{data augmentation}
As pointed out by \citet{chen2020simple}, the selection of a data augmentation strategy is crucial to the performance of contrastive learning. 
After considering several augmentation methods such as synonym replacement, inspired by \citet{lowell2020unsupervised}, we apply a simple yet useful random token substitution-based method to augment the original examples (see Appendix~\ref{appendix:data augmentation} for details of our selection).  

More concretely, we first build a vocabulary\footnote{Details are described in Appendix~\ref{appendix:config:tok}.} $V$ for the given training corpus, then apply an operation $T$ to transform each token $w_{i}$ in the input sentence $x_{i} = (w_{1}, w_{2}, \ldots, w_{k})$ to $T(w_{i})$:
\begin{linenomath}
\begin{equation}
\label{eq:data augmentation}
T(w_{i})=\left\{
\begin{aligned}
& w_{i}\quad \mbox{with probability $p$} \\
& w^{\prime}_{i}\quad \mbox{with probability $1-p$},
\end{aligned}
\right.
\end{equation}
\end{linenomath}
where $w^{\prime}_{i}$ is a random token extracted from $V$ and $p \in [0,1]$ controls the diversity of the augmentations. Although this method cannot ensure the fluency of the generated sequence $T(x_{i})$, it enables our model to learn adequate instance-wise features from the raw text and hence is useful for generating high-quality representations.

\subsection{Label-Free Adversarial Example Generator (Adv-Generator)}
\label{adv-gen}
Since SCAT needs to generate adversarial examples 
without using ground-truth labels, conventional state-of-the-art attackers cannot be applied here directly. Instead, we 
design a method to generate label-free adversarial examples for pre-training efficiently.   
The whole generation process is summarized in Algorithm~\ref{alg:1}, which can be divided into two main steps to be elaborated below.

\noindent \textbf{Determine Attack Positions (line 1-8).} \quad Given two augmentations of an input sentence $x_{i}$, similar to \citet{kim2020adversarial}, we build $x_{i}$'s label-free adversarial counterpart $x_{i}^{adv}$ upon the copy of one of the augmented examples $T(x_{i})_{1}$, named $T(x_{i})^{c}_{1}$. 
However, without gold labels, we could neither decide the attack positions via calculating the token importance scores like \citet{jin2020bert}, nor end the attack until the model's prediction has been changed. To determine the attack positions, we adopt the following steps: for each token $w_{j}$ in $T(x_{i})^{c}_{1}$, let $\nabla_{w_{j}}L$ be the gradient of the contrastive loss with respect to the embedding vector of $w_{j}$, where $L$ is the contrastive loss between two augmentations $T(x_{i})_{1}$ and $T(x_{i})_{2}$, calculated by Eq.\ref{eq1} and Eq.\ref{eq2}. We then define the importance score $I_{w_{j}}$ for $w_{j}$ as the 1-norm of the gradient vector $\nabla_{w_{j}}L$, i.e., $I_{w_{j}} = \Vert \nabla_{w_{j}}L \Vert_{1}$. A larger importance score $I_{w_{j}}$ for $w_{j}$ indicates a more signiﬁcant contribution of the corresponding token to the change in the contrastive loss. \sref{ablation} demonstrates the effectiveness of this important score ranking method. The tokens with the largest importance scores will be identified as the attack positions.

Next, we rank the words in $T(x_{i})^{c}_{1}$ based on their importance scores in descending order and build an attack set $\mathit{ATT}$. Note that for each sequence, we steadily pick the $\epsilon$ most important words to form $\mathit{ATT}$ since we lack ground-truth labels to determine when to stop the attack. 

\noindent \textbf{Generate Label-Free Adversarial Example (line 9-20).}\quad
Recent works highly praised pre-trained mask language models in textual attacks, due to their ability to generate fluent and semantically consistent substitutions for input sentences \cite{li2020bert, li2020contextualized}. Following \citet{li2020bert}, we first use Bytes-Pair-Encoding (BPE) to tokenize $T(x_{i})^{c}_{1}$ into sub-word tokens $T(x_{i})^{sub}_{1} = (s_{1}, s_{2}, ..., s_{m})$ and thus obtain $\mathit{ATT}_{sub}$, the sub-word tokenized counterpart of $\mathit{ATT}$. Since the tokens in $\mathit{ATT}_{sub}$ are selected from $T(x_{i})^{sub}_{1}$, we can yield $K$ most possible substitutions for each token in $\mathit{ATT}_{sub}$ by feeding $T(x_{i})^{sub}_{1}$ into a pre-trained BERT 
model. 

Afterwards, we start to replace tokens. Given a target word $w_{p}$ in $\mathit{ATT}$, if it is tokenized into sub-words in $\mathit{ATT}_{sub}$, we first calculate the perplexity scores of possible sub-word combinations, then rank these perplexity scores to extract the top-$K$ combinations, which form a substitution candidate set $C$. Otherwise, substitutions generated in line~9 will consist of $C$. A synonym dictionary \cite{mrkvsic2016counter} is then applied to exclude antonyms from $C$ since masked language models could not discriminate synonyms from antonyms. We do not filter out stop words to keep consistent with our data augmentation method (\sref{data augmentation}), where stop words also act as potential substitutions.

Lastly, instead of iterating all the items, we randomly pick one item from the candidate set $C$ to replace the target word $w_{p}$. This strategy is a trade-off between attack efficiency and effectiveness, since searching for optimal adversarial examples is time-consuming and thus impractical for pre-training. Nevertheless, as demonstrated by the experimental results in \sref{main}, our attack strategy can still generate representative label-free adversarial examples for SCAT to learn robust representations. We also tested the effectiveness of our label-free attack and results are mentioned in Appendix~\ref{appendix:attack performance}. After going through all the words in $\mathit{ATT}$, we obtain the final label-free adversarial example $x_{i}^{adv}$. In addition, we take the contrastive loss between $x_{i}^{adv}$ and the other clean augmentation as an extra regularizer in the final training objective (Eq.\ref{eq7}).

\subsection{Learning Objective}
\label{objective}
We can now formulate our learning objective. Specifically, we add $x_{i}^{adv}$ to the original positive set $\lbrace x_{i_{pos}} \rbrace$ so the final training set $B$ will include 3$N$ examples due to the expansion of $\lbrace x_{i_{pos}} \rbrace$. Then we learn robust representations by minimizing the contrastive loss between two augmentations and $x_{i}^{adv}$. Using the same settings in \sref{pre}, we first define the contrastive objective with three samples in $\lbrace x_{i_{pos}} \rbrace$ following Eq.\ref{eq1} and Eq.\ref{eq2}:
\begin{linenomath}
\begin{equation}
\begin{aligned}
    L_{\text{CL}_{3}( T(x_{i})_{1}, T(x_{i})_{2}, x_{i}^{adv})} &= L_{\hat{\text{CL}}(T(x_{i})_{1}, T(x_{i})_{2})}\\
    &+ L_{\hat{\text{CL}}(T(x_{i})_{1}, x_{i}^{adv})}\\
    &+ L_{\hat{\text{CL}}(T(x_{i})_{2}, x_{i}^{adv})}
\end{aligned}
\end{equation}
\end{linenomath}
Note that $L_{\hat{\text{CL}}}$ is a modification of $L_{\text{CL}}$ while the size of the set $B$ is increased to $3N$ in Eq.\ref{eq1}. The final objective for the pre-training part can thus be calculated as:
\begin{linenomath}
\begin{align}
\label{eq7}
    L_{\text{CL}_{adv}} &= L_{\text{CL}_{3}(T(x_{i})_{1}, T(x_{i})_{2}, x_{i}^{adv})}\\ 
    L_{\text{Reg}} &= L_{\text{CL}(x_{i}^{adv}, T(x_{i})_{2})}\\
    L_{\text{Final}} &= L_{\text{CL}_{adv}} + \lambda L_{\text{Reg}},
\end{align}
\end{linenomath}
where $\lbrace T(x_{i})_{1}, T(x_{i})_{2}, x_{i}^{adv} \rbrace$ forms the new positive set, augmentations of the other instances in the same mini-batch correspond to the negative set, and $\lambda$ is used to control the regularization strength. 

\subsection{Linear Evaluation}
Since pre-trained representations cannot be used for text classification directly, we follow the existing self-supervised learning method (\cite{chen2020simple}) to leverage an MLP layer at the top of the encoder $E$'s outputs to predict the labels for different input examples. We train this MLP layer on the original training set in a supervised manner while fixing all the parameters of $E$. The cross-entropy loss is adopted as the optimization objective here. While this simple, efficient linear evaluation method helps SCAT achieve promising robustness (as shown in Table~\ref{tab:main results}), we also analyzed another conventional evaluation strategy: fine-tuning the whole model. Details can be found in Appendix~\ref{appendix:evaluation}.

\section{Experiments}
\label{exp}
\subsection{Setup}
\noindent \textbf{Datasets.} \quad We evaluate our method on two representative text classification datasets: AG's News (AG) and DBPedia \cite{zhang2015character}. Detailed description and statistics of these two datasets are included in Appendix~\ref{appendix:stats}. For AG, we perform experiments on the 1k test examples collected by \citet{jin2020bert} since their data splits have been widely used. For DBPedia, we randomly selected 1k samples from its original test set for experiments. Since both AG and DBPedia do not provide a validation set, we randomly picked 2k instances from their original test sets for validation, while not overlapping with the 1k test examples.

\label{configuration}
\noindent \textbf{Configuration.} \quad During pre-training, we adopt two backbone encoders: 1) base sized BERT (BERT$_{base}$, \citet{devlin2019bert}); 2) randomly initialized Transformer \cite{vaswani2017attention} with the same architecture as BERT (12 layers, 12 heads, and hidden layer size of 768). 
More implementation details regarding the Adv-Generator, projector, optimization, linear evaluation, and data pre-processing are described in Appendix~\ref{appendix:config}.

%

\noindent \textbf{Attackers.} \quad In our experiments, we use two state-of-the-art attackers to evaluate model robustness: TextFooler and BERT-Attack. See Appendix~\ref{appendix:attacker} for detailed descriptions of these two attackers. We implement the two attackers following their publicly released versions. For TextFooler, the thresholds of both synonym and sentence similarity scores are set to 0.5
, according to its authors' instructions. As for BERT-Attack, 
on AG, \citet{li2020bert} report results without filtering out antonyms from the potential substitutions. We instead do extra evaluations on both datasets with the antonym filtering process for fair comparison, since masked language models could not distinguish synonyms from antonyms. 

\noindent \textbf{Evaluation Metrics.} \quad To measure the robustness of SCAT from different perspectives, we introduce several evaluation metrics: (1)~\textbf{Clean Accuracy (Acc):} model's accuracy score on clean examples; (2)~\textbf{After-Attack Accuracy (Atk Acc):} model's accuracy score after being attacked; (3)~\textbf{Attack Failure Rate (AFR):} percentage of adversarial examples that fail to change the model's prediction.
For all the metrics especially the last two core measures, higher values indicate better performance. 


\begin{table*}[tb]
\renewcommand\arraystretch{1.1}
  \centering
   \setlength{\tabcolsep}{1.5mm}{
    \small
    \begin{tabular}{cl|*3{c}|*3{c}}
    \toprule[1pt]
    \multicolumn{2}{c|}{} & \multicolumn{3}{c|}{\textbf{AG}} & \multicolumn{3}{c}{\textbf{DBPedia}} \\
      \textbf{Attacker} & \textbf{Model} & \textbf{Acc} & \textbf{Atk Acc} & \textbf{AFR} 
      & \textbf{Acc} & \textbf{Atk Acc} & \textbf{AFR} 
      \\
     \midrule[1pt]
     \multirow{6}*{TextFooler} & Transformer & 92.3 & 0.3 & 0.3 
     & 98.7 & 8.7 & 8.1 
     \\
     ~ & CL-Transformer & 87.7 & 14.5 & 16.5 
     & 93.7 & 12.9 & 13.8 
     \\
     ~ & SCAT-Transformer & 90.4 & \textbf{25.4} & \textbf{28.1} 
     & 92.9 & \textbf{17.9} & \textbf{19.3} 
     \\
     \cline{2-8}
     ~ & BERT & 95.3 & 16.6 & 17.4 
     & 99.4 & 20.1 & 20.2 
     \\
     ~ & CL-BERT & 92.5 & 38.1 & 41.2 
     & 98.9 & 32.5 & 32.8 
     \\
     ~ & SCAT-BERT & 92.2 & \textbf{45.1} & \textbf{49.0} 
     & 98.5 & \textbf{48.3} & \textbf{49.0}
     \\
     \midrule[1pt]
     \multirow{6}*{BERT-Attack} & Transformer  & 92.3 & 0.5/3.4 & 0.5/3.7 
     & 98.7 & 0.6/4.6 & 0.6/4.7 
     \\
     ~ & CL-Transformer & 87.7 & 13.4/25.5 & 15.2/29.0 
     & 93.7 & 9.0/23.4 & 9.6/24.9 
     \\
     ~ & SCAT-Transformer & 90.4 & \textbf{17.9}/\textbf{44.0} & \textbf{19.8}/\textbf{48.6} 
     & 92.9 & \textbf{10.4}/\textbf{45.7} & \textbf{11.2}/\textbf{49.1} 
     \\
     \cline{2-8}
     ~ & BERT & 95.3 & 19.8/41.7 & 20.8/43.8 
     & 99.4 & 16.8/46.1 & 16.9/46.4 
     \\
     ~ & CL-BERT & 92.5 & \textbf{36.8}/52.2 & \textbf{39.8}/56.4 
     & 98.9 & 24.5/56.3 & 24.8/57.0 
     \\
     ~ & SCAT-BERT & 92.2 & 31.7/\textbf{58.7} & 34.5/\textbf{63.7} 
     & 98.5 & \textbf{27.2}/\textbf{74.3} & \textbf{27.7}/\textbf{75.5} 
     \\
     \bottomrule[1pt]
    \end{tabular}
  }    
    \caption{Experimental results of different models against TextFooler and BERT-Attack on AG and DBPedia. For evaluation metrics except for the clean accuracy (Acc), the best performance of models under each attack is boldfaced. On both datasets, we test BERT-Attack without/with antonym filtering and the results are separated correspondingly. Except for supervised Transformer and BERT, metric values of models are averaged over three different runs. We also provide complete results in Appendix~\ref{appendix:complete main} for reference.}
  \label{tab:main results}%
\end{table*}

\noindent \textbf{Baseline Models.} \quad For each backbone encoder mentioned in \sref{configuration}, we perform experiments with two different baselines. Taking BERT as an example, these baselines are (1)~\textbf{BERT:} a base model trained with clean examples in a supervised manner; (2)~\textbf{CL-BERT:} a self-supervised contrastive learning-based model pre-trained without using extra adversarial examples. To reduce the effect of randomness brought by data augmentation and label-free adversarial example generation during pre-training, models including these steps were run with three different random seeds and we report the average metric scores in all the experiments.

\subsection{Main Results}
\label{main}
\textbf{SCAT Improves Robustness.} \quad \tref{tab:main results} summarizes main results on the two datasets. As we can see, supervised models encounter serious performance drop against the two attackers. By contrast, SCAT improves the robustness of these standard models, demonstrated by the significantly higher results across the board. For both datasets, SCAT-BERT and SCAT-Transformer outperform BERT and Transformer by an average attack failure rate of 24.9\% against TextFooler and 24.1\% against BERT-Attack.\footnote{The calculation process is detailed in Appendix~\ref{appendix:calculation}.}
Since the key idea behind the two attackers differs, these impressive results illustrate the possibility to train a robust model that can defend against different types of attacks. It is worth noting that SCAT does not sacrifice the clean accuracy much compared to the supervised models, and the average drop is only 2.9\%. This result is surprising since SCAT’s pre-training stage is fully label-free and it further
suggests that SCAT can still generate high-quality representations for downstream tasks while largely improving model robustness.

\noindent \textbf{Comparison between SCAT and CL.} \quad  Interestingly, we observe that using contrastive learning alone usually enhances model robustness, 
which proves that our data augmentation method helps models adapt to potential perturbations to some certain extent. Nevertheless, compared to CL, SCAT has more benefits. First, as shown in \tref{tab:main results}, SCAT has a significantly stronger robustness performance. For example, for the two robustness-related metrics, SCAT models outperform CL models by 10.9\%, 11.6\% on AG and 5.0\%, 15.8\% on DBPedia respectively against TextFooler.

Moreover, SCAT can defend against high-quality adversarial examples better, demonstrated by its higher metric scores against the attackers with strict semantic constraints like TextFooler and BERT-Attack with antonym filtering. As shown in Table~\ref{tab:main results}, SCAT enhances Transformer and BERT's performance with respect to the after-attack accuracy by an average of 31.7\% against BERT-Attack with antonym filtering.\footnote{The calculation process is detailed in Appendix~\ref{appendix:calculation}.}
Although CL-BERT outperforms SCAT-BERT when skipping the antonym filtering process under BERT-Attack on AG, adversarial examples obtained here may not be fluent and could be easily identified by heuristic rules due to the potential misuse of antonyms during token substitution. In contrast, for high-quality adversarial examples that are hard to be detected in advance, SCAT-BERT can defend them better than CL-BERT due to the robust pre-training stage.


\noindent \textbf{Results on Pre-trained Encoder.} \quad Another advantage of SCAT is that except for training robust models from scratch, it can also perform robust fine-tuning for pre-trained language models without modifying their structures. For example, SCAT-BERT enhances the after-attack accuracy of BERT by 28.5\% and 28.2\% on both datasets against TextFooler, while only decreasing 3.1\% and 0.9\% of the clean accuracy. These results further indicate that SCAT fits well for the current trend of fine-tuning huge pre-trained language models.

\begin{table}[tb]
\renewcommand\arraystretch{1.1}
  \centering
    \resizebox{0.47\textwidth}{!}{
    \small
    \begin{tabular}{lc|*2{c}|*2{c}}
    \toprule[1pt]
     & & \multicolumn{2}{c|}{\textbf{TextFooler}} & \multicolumn{2}{c}{\textbf{BERT-Attack}} \\
       \textbf{Model} & \textbf{Acc} & \textbf{Atk Acc} & \textbf{AFR} 
       & \textbf{Atk Acc} & \textbf{AFR} 
      \\
     \midrule[1pt]
      Transformer & 92.3  & 0.3 & 0.3 
      & 0.5/3.4 & 0.5/3.7 
      \\
     +CL+Extra Aug & 88.7 & 19.7 & 22.2 
     & 17.7/31.2 & \textbf{19.9}/35.1 
     \\
     +SCAT & 90.4 & \textbf{25.4} & \textbf{28.1} 
     & \textbf{17.9}/\textbf{44.0} & 19.8/\textbf{48.6} 
     \\
     \midrule[1pt]
     BERT & 95.3  & 16.6 & 17.4 
      & 19.8/41.7 & 20.8/43.8 
      \\
     +CL+Extra Aug & 92.7 & 44.4 & 48.0 
     & \textbf{41.9}/55.9 & \textbf{45.2}/60.3 
     \\
     +SCAT & 92.2 & \textbf{45.1} & \textbf{49.0} 
     & 31.7/\textbf{58.7} & 34.5/\textbf{63.7} 
     \\
     \bottomrule[1pt]
    \end{tabular}
  }
    \caption{Effect of positive set size. We use ``+" here to distinguish different models for brevity. Except for the supervised Transformer and BERT, the metric values of models are averaged over three runs. Complete results are shown in Appendix~\ref{appendix:complete ablation}.}
  \label{tab:set size}
\end{table}

\begin{table}[tb]
\renewcommand\arraystretch{1.1}
  \centering
    \resizebox{0.47\textwidth}{!}{
    \small
    \begin{tabular}{lc|*2{c}|*2{c}}
    \toprule[1pt]
     & & \multicolumn{2}{c|}{\textbf{TextFooler}} & \multicolumn{2}{c}{\textbf{BERT-Attack}} \\
       \textbf{Model} & \textbf{Acc} & \textbf{Atk Acc} & \textbf{AFR} 
       & \textbf{Atk Acc} & \textbf{AFR} 
      \\
     \midrule[1pt]
      Transformer & 92.3  & 0.3 & 0.3 
      & 0.5/3.4 & 0.5/3.7 
      \\
     +SCAT(Random) & 91.5 & 22.0 & 24.0 
     & \textbf{18.0}/42.7 & 19.6/46.6 
     \\
     +SCAT & 90.4 & \textbf{25.4} & \textbf{28.1} 
     & 17.9/\textbf{44.0} & \textbf{19.8}/\textbf{48.6} 
     \\
     \midrule[1pt]
     BERT & 95.3  & 16.6 & 17.4 
      & 19.8/41.7 & 20.8/43.8 
      \\
     +SCAT(Random) & 92.1 & 41.5 & 45.0 
     & \textbf{37.5}/\textbf{61.9} & \textbf{40.7}/\textbf{67.2} 
     \\
     +SCAT & 92.2 & \textbf{45.1} & \textbf{49.0} 
     & 31.7/58.7 & 34.5/63.7 
     \\
     \bottomrule[1pt]
    \end{tabular}
  }
    \caption{Effect of gradient-based ranking, with similar settings as Table \ref{tab:set size}.}
  \label{tab:gradient ranking}
\end{table}

\subsection{Ablation Study}
\label{ablation}
To assess the effectiveness of different modules in SCAT, we conduct an ablation study on AG with two additional baseline models (for each encoder). The first model does not utilize the Adv-Generator. Instead, it creates an extra augmentation for a given example and adds this extra sequence to the positive set as the label-free adversarial example. The other model removes the gradient-based ranking step in Algorithm~\ref{alg:1} and determines the attack positions randomly. Table \ref{tab:set size}, \ref{tab:gradient ranking} show the results. Since the comparison between SCAT, CL and supervised models has been discussed 
in \sref{main}, we focus on comparing SCAT with the two new baselines here.

\noindent \textbf{Effect of Ppositive Sset Ssize.} \quad We first compare SCAT against the baseline (CL+Extra Aug) that contrasts three random augmentations. As can be seen in \tref{tab:set size}, SCAT nearly outperforms this baseline across the 
board, indicating that label-free adversarial examples generated by the Adv-Generator (\sref{adv-gen}) do highly contribute to the improvement of model robustness. The only exception appears when defending BERT-Attack without antonym filtering, which is not surprising since this baseline is an augmented version of CL. As mentioned in \sref{main}, CL-based models are more robust to lower-quality adversarial examples that can be easily identified, since their representations are merely learned from random augmentations and should be more familiar with unnatural token substitutions.

\noindent \textbf{Effect of Ggradient-Bbased Rranking.} \quad We now switch to the ranking method in the Adv-Generator. As shown in \tref{tab:gradient ranking}, replacing the ranking process with random selection (SCAT(Random) leads to a drop in robustness 
against TextFooler. As for BERT-Attack, SCAT outperforms SCAT(Random) with Transformer, while SCAT(Random) has better robustness results for the BERT model, probably because attack positions determined by the gradient of contrastive loss are more consistent with those used in TextFooler. While assigning attack positions, TextFooler separately considers the situation if deleting one token from the original input changes the target model's prediction compared to BERT-Attack. Overall, these results justify our choice of applying the gradient-based ranking strategy in the Adv-Generator. 

\begin{table}[tb]
\renewcommand\arraystretch{1.1}
  \centering
    \resizebox{0.47\textwidth}{!}{
    \small
    \begin{tabular}{cl|*3{c}}
    \toprule[1pt]
      \textbf{Attacker} & \textbf{Model} & \textbf{Acc} & \textbf{Atk Acc} & \textbf{AFR} 
      \\
     \midrule[1pt]
     \multirow{4}*{\makecell[c]{Text-\\Fooler}} & BERT & 95.3  & 16.6 
     & 17.4  \\
     ~ & +Adv & 94.7 & 29.0 
     & 30.6 \\
     ~ & SCAT-BERT(s1) & 91.7 & 43.8 
     & 47.8 \\
     ~ & +Adv & 92.1  & \textbf{50.0} 
     & \textbf{54.3} \\
     \midrule[1pt]
     \multirow{4}*{\makecell[c]{BERT-\\Attack}} & BERT & 95.3  & 19.8/41.7 
     & 20.8/43.8  \\
     ~ & +Adv & 94.3/94.3 & 29.8/51.8 
     & 31.6/54.9 \\
     ~ & SCAT-BERT(s1) & 91.7 & 31.5/57.9 
     & 34.4/63.1  \\
     ~ & +Adv & 91.5/92.1  & \textbf{40.0}/\textbf{62.2} 
     & \textbf{43.7}/\textbf{67.5}\\
     \bottomrule[1pt]
    \end{tabular}
  }    
    \caption{Supervised adversarial training results on AG. 
    For SCAT-BERT, we take one of its three runs (s1) for evaluation. For each ``+Adv'' model under BERT-Attack, we trained two versions without/with antonym filtering and 
    tested them in corresponding situations.}
  \label{tab:flexibility}
\end{table}

\subsection{Combining SCAT with Supervised Adversarial Training}
\label{sup-adv}
Existing works \cite{jin2020bert, li2020bert} show that adversarial examples generated by attackers can be used to enhance model robustness via supervised adversarial training. 
In this part, we further evaluate the flexibility of SCAT by training it with labeled adversarial examples. Specifically, we follow \citet{jin2020bert} to expand AG's training set using labeled adversarial examples crafted by TextFooler and BERT-Attack. Since it will be time-consuming to attack a specific victim model on the whole training set, we only pick BERT and one run of SCAT-BERT for comparison. The moderate standard deviation scores for different runs of SCAT-BERT in \tref{tab:mean} justify this setting. 
When testing BERT, we fine-tune a new BERT from scratch on the expanded dataset. For SCAT-BERT, we perform linear evaluation on the expanded dataset while fixing the pre-trained encoder.

\tref{tab:flexibility} lists the results on AG's test set used earlier. As can be seen, performing adversarial training directly using the generated adversarial examples does help BERT defend against attacks better. However, the performance gap between BERT+Adv and SCAT-BERT is still significant, confirming the effectiveness of our method. Notably, adding labeled adversarial examples to the linear evaluation part further boosts both the robustness scores and the clean accuracy of SCAT-BERT, with an average improvement of 6.3\% on the after-attack accuracy and 6.7\% on the attack failure rate among all attack types. Since simply training a linear classifier might not make full use of the extra information, other potential methods such as pre-training on the expanded dataset using SCAT should further improve the results. This again illustrates the flexibility of SCAT and largely widens its application scope.

\section{Conclusion}
In this work, we propose a novel adversarial training framework named SCAT, which can learn robust textual representations in a fully label-free manner. We first come up with a token substitution-based method to craft adversarial examples from augmentations of the data without requiring ground-truth labels. Next, we implement adversarial training via minimizing the contrastive loss between the augmentations and their adversarial counterparts. In the experiments, we adopt two state-of-the-art attack algorithms on two text classification datasets to evaluate the robustness of different models. Our experimental results demonstrate that SCAT can both train robust language models from scratch and improve the robustness of pre-trained language models significantly. Moreover, we show the effectiveness of different modules of SCAT through an ablation study. Finally, we illustrate the flexibility of SCAT by combining it with supervised adversarial training, motivating further research in this area.

\section{Ethical Considerations}
Our proposed SCAT framework can be used to improve the robustness of existing text classification systems. Since SCAT is very effective and flexible, it can be widely used in the NLP community. Nevertheless, like other defense methods, SCAT can lead to a slight decrease in the clean accuracy for classification tasks. This trade-off between accuracy and robustness has to be taken into consideration in the context of the specific application when trying to decide whether to incorporate a defense scheme such as SCAT.

\bibliography{main}
\bibliographystyle{acl_natbib}

\appendix

\section{Adversarial Training}
\label{appendix:adversarial training}
We first introduce the key idea of adversarial training here. For continuous-valued image data, \citet{madry2018towards} proposed to perform adversarial training via solving a min-max optimization problem:
\begin{linenomath}
\begin{equation}
    \underset{\theta}{{\rm min}} \, \mathbbm{E}_{(x,y)\sim\mathbbm{D}} \left[\underset{\|\delta\|_{\infty}\leq\epsilon}{{\rm max}} l(\theta, x+\delta, y)\right]
\end{equation}
\end{linenomath}
where $\theta$, $\delta$, and $l$ denote the model parameters, projected perturbations and loss function, respectively, and $(x, y)$ is a labeled instance from the training set $\mathbbm{D}$. When turning to the text domain, existing works usually follow this optimization scheme with discrete adversarial examples. 

\begin{table}[tb]
\renewcommand\arraystretch{1.1}
  \centering
    \resizebox{0.47\textwidth}{!}{
    \small
    \begin{tabular}{cl|*3{c}}
    \toprule[1pt]
      \textbf{Attacker} & \textbf{Model} & \textbf{Acc} & \textbf{Atk Acc} & \textbf{AFR} 
      \\
     \midrule[1pt]
     \multirow{4}*{\makecell[c]{Text-\\Fooler}} & CL-Transformer(s1) & 84.6  & 13.1 
     & 15.5  \\
     ~ & Synonym-CL-Transformer & 34.0 & 25.3 
     & 74.4 \\
     ~ & CL-BERT & 92.4 & 37.2 
     & 40.3 \\
     ~ & Synonym-CL-BERT & 89.4 & 63.4 
     & 70.9 \\
     \midrule[1pt]
     \multirow{4}*{\makecell[c]{BERT-\\Attack}} & CL-Transformer & 84.6  & 11.9/21.7 
     & 14.1/25.7  \\
     ~ & Synonym-CL-Transformer & 34.0 & 1.6/1.6 
     & 4.7/4.7 \\
     ~ & CL-BERT & 92.4 & 33.6/50.3 
     & 36.4/54.4 \\
     ~ & Synonym-CL-BERT & 89.4 & 18.6/30.2 
     & 20.8/33.8 \\
     \bottomrule[1pt]
    \end{tabular}
  }
    \caption{Comparison of two data augmentation methods on AG. We picked one run of CL-Transformer and CL-BERT for comparison here, and use the same random seed to train synonym augmented models.}
  \label{tab:comparison of data augmentation}
\end{table}

\section{Comparison between Different Data Augmentation Methods}
\label{appendix:data augmentation}
 In our preliminary experiments on AG, we considered two types of augmentation methods. The first type is the strategy in \sref{data augmentation}. The other type is synonym substitution, where a synonym set for each token in vocabulary $V$ is extracted from a synonym dictionary following steps and configurations in \citet{jin2020bert}. For each token in the input sentence, we also transform it using Eq.\ref{eq:data augmentation} while $w^{\prime}_{i}$ is picked from the token's synonym set. If the token does not appear in the synonym dictionary, we will not replace this token. For both augmentation methods, we perform contrastive pre-training and linear evaluation and use two attackers to attack them like \sref{main}. 
 
 Table \ref{tab:comparison of data augmentation} visualizes the results of our preliminary experiments. We observe that synonym substitution-based augmentations could not lead to good representations compared to random token substitution, especially when pre-training a transformer encoder from scratch. As can be seen, Synonym-CL-Transformer only achieves 34.0\% clean accuracy scores. This may be because although we can generate two semantic coherent augmentations for contrastive learning using synonym substitution, the high-dimensional representations of these two augmentations are too similar for our model to learn enough linguistic features from the given input sentence. Moreover, Synonym-CL based models are more likely to overfit the synonym set they used in pre-training, which will seriously hurt their generalization abilities. As shown in Table \ref{tab:comparison of data augmentation}, Although Synonym-CL-Transformer and Synonym-CL-BERT have strong robustness performances against TextFooler, their abilities to defend BERT-Attack are quite weak. On the contrary, CL-Transformer and CL-BERT have moderate scores against different types of attack. Based on the above reasons, we choose random token substitution as our data augmentation method in this work. But we do agree that random token substitution might not be the most suitable strategy and regard finding a better data augmentation method as a future research direction.

\section{Further Details on the Configuration}
\label{appendix:config}

\subsection{Use of Different Tokenizers}
\label{appendix:config:tok}
In this work, we adopt two tokenizers for different purposes. The first one is the NLTK\footnote{\url{https://www.nltk.org/}} word tokenizer, which is used to tokenize sentences in a given dataset to build the vocabulary $V$ in \sref{data augmentation}. It is also adopted to tokenize a given sentence before the token substitution step in Adv-Generator starts. The BERT tokenizer provided by Hugging Face\footnote{\url{https://huggingface.co/transformers}} (uncased version) is used in the remaining tokenization scenes. The maximum sequence length is set to 128 for both datasets.

\subsection{Pre-Training}
\label{appendix:config:pre}
Recall that during pre-training, we adopt two different backbone encoders: 1) base sized BERT (BERT$_{base}$); 2) randomly initialized Transformer with the same architecture as BERT (12 layers, 12 heads, hidden layer size of 768). During data augmentation, the replacing probability $p$ is set to 0.3. Regarding our Adv-Generator, we set $\epsilon$ to 30\% 
and the number of candidates $K$ to 48. For the projector, we take a 2-layer MLP module to project the encoder's output to a 128-dimensional hidden space. As for optimization, we use AdamW with a learning rate of 5e-5 and weight decay of 1e-6. In the final loss, we adopt 0.5 for the temperature parameter $\tau$ and $1/256$ for $\lambda$. On AG, each model is pre-trained for 50 epochs with a batch size of 32, while the first 3 epochs act as the warm-up session. On DBPedia, since its training set is about 5 times larger than AG, to be consistent on the overall training steps, each model is pre-trained for 10 epochs while the first epoch acts as the warm-up session. For supervised baselines trained on clean examples, the number of warm-up epochs is set to 0. The batch size is selected from $\lbrace 16, 32, 64, 128 \rbrace$ based on the robustness performance. The three random seed numbers are 2020, 2010, and 2000 respectively. SCAT takes about 6 days to train 50 epochs on AG and 10 epochs on DBPedia using one RTX 3090 GPU. All the codes are implemented with PyTorch. \footnote{\url{https://pytorch.org/}}

\subsection{Linear Evaluation}
\label{appendix:config:lin}
At the linear evaluation stage, we train the 1-layer linear classifier on top of the encoder $E$'s outputs for 100 epochs with a batch size of 128, then pick the epoch that performs best on the validation set for the final usage. Optimization is done using AdamW with a value of 5e-4 for both the learning rate and weight decay parameters.

\begin{table}[tb]
\renewcommand\arraystretch{1.1}
  \centering
    \resizebox{0.47\textwidth}{!}{
    \small
    \begin{tabular}{cl|*3{c}}
    \toprule[1pt]
      \textbf{Dataset} & \textbf{Model} & \textbf{Acc} & \textbf{Atk Acc} & \textbf{AFR} 
      \\
     \midrule[1pt]
     \multirow{2}*{\makecell[c]{AG}} & Transformer & 92.3  & 73.5 
     & 79.6  \\
     ~ & BERT & 95.3 & 85.0 
     & 89.2 \\
     \midrule[1pt]
     \multirow{2}*{\makecell[c]{DBPedia}} & Transformer & 98.7  & 88.1 
     & 89.3  \\
     ~ & BERT & 99.4 & 92.9 
     & 93.5 \\
     \bottomrule[1pt]
    \end{tabular}
  }    
    \caption{Label-free attack results.}
  \label{tab:label-free attack results}
\end{table}

\section{Performance of our Label-Free Attack}
\label{appendix:attack performance}
In \sref{adv-gen}, we design a label-free method to dynamically craft adversarial examples during pre-training using the gradient of contrastive loss. In this section, we provide a more comprehensive view of our proposed attack by evaluating it with supervised Transformer and BERT. Results are listed in Table \ref{tab:label-free attack results}. 

As can be seen, our attack successfully hurt supervised models' robustness to a certain extent. Although the power of our label-free attack is not as strong as state-of-the-art attacking methods, as mentioned in \sref{adv-gen}, this is a trade-off between attack efficiency and effectiveness due to our limited computational resources. Nevertheless, these label-free adversarial examples still highly improve model robustness as illustrated in \sref{exp}.

\begin{table*}[tb]
\renewcommand\arraystretch{1.1}
  \centering
    \small
    \begin{tabular}{cl|*3{c}|*3{c}}
    \toprule[1pt]
    \multicolumn{2}{c|}{} & \multicolumn{3}{c|}{\textbf{AG}} & \multicolumn{3}{c}{\textbf{DBPedia}} \\
      \textbf{Attacker} & \textbf{Model} & \textbf{Acc} & \textbf{Atk Acc} & \textbf{AFR} 
      & \textbf{Acc} & \textbf{Atk Acc} & \textbf{AFR} 
      \\
     \midrule[1pt]
     \multirow{6}*{TextFooler} & Transformer & 92.3 & 0.3 & 0.3 
     & 98.7 & 8.7 & 8.1 
     \\
     ~ & SCAT-Transformer(s1)(FT) & 94.0 & 19.1 & 20.3 
     & 99.2 & 12.7 & 12.8 
     \\
     ~ & SCAT-Transformer(s1) & 90.4 & \textbf{26.3} & \textbf{29.1} 
     & 94.0 & \textbf{17.3} & \textbf{18.4} 
     \\
     \cline{2-8}
     ~ & BERT & 95.3 & 16.6 & 17.4 
     & 99.4 & 20.1 & 20.2 
     \\
     ~ & SCAT-BERT(s1)(FT) & 95.4 & 21.5 & 22.5 
     & 99.4 & 36.7 & 36.9 
     \\
     ~ & SCAT-BERT(s1) & 91.7 & \textbf{43.8} & \textbf{47.8} 
     & 98.3 & \textbf{48.5} & \textbf{49.3}
     \\
     \midrule[1pt]
     \multirow{6}*{BERT-Attack} & Transformer  & 92.3 & 0.5/3.4 & 0.5/3.7 
     & 98.7 & 0.6/4.6 & 0.6/4.7 
     \\
     ~ & SCAT-Transformer(s1)(FT) & 94.0 & 8.1/30.9 & 8.6/32.9 
     & 99.2 & 5.4/34.9 & 5.4/35.2 
     \\
     ~ & SCAT-Transformer(s1) & 90.4 & \textbf{19.1}/\textbf{44.6} & \textbf{21.1}/\textbf{49.3} 
     & 94.0 & \textbf{9.3}/\textbf{49.4} & \textbf{9.9}/\textbf{52.6} 
     \\
     \cline{2-8}
     ~ & BERT & 95.3 & 19.8/41.7 & 20.8/43.8 
     & 99.4 & 16.8/46.1 & 16.9/46.4 
     \\
     ~ & SCAT-BERT(s1)(FT) & 95.4 & 9.7/33.7 & 10.2/35.3 
     & 99.4 & 21.2/63.2 & 21.3/63.6 
     \\
     ~ & SCAT-BERT(s1) & 91.7 & \textbf{31.5}/\textbf{57.9} & \textbf{34.4}/\textbf{63.1} 
     & 98.3 & \textbf{27.1}/\textbf{75.1} & \textbf{27.6}/\textbf{76.4} 
     \\
     \bottomrule[1pt]
    \end{tabular}
    \caption{Comparison of the two evaluation methods. FT means fine-tune the pre-trained model for downstream tasks.}
  \label{tab:two evaluation methods}
\end{table*}

\section{Analysis of Evaluation Methods for SCAT}
\label{appendix:evaluation}
In our experiments, we train a linear layer on top of the fixed SCAT encoder to perform linear evaluation for downstream tasks, following the standard setting of existing self-supervised learning models. While this simple method achieves great performance as illustrated in \sref{exp}, there also exists another popular evaluation protocol in the NLP community, which is to fine-tune the whole model for the downstream task. In this section, we test the robustness of these two evaluation schemes against two attackers. Similar to \sref{sup-adv}, we only pick one run of SCAT-Transformer and SCAT-BERT for comparison. For the fine-tuning method, we fine-tune the pre-trained SCAT encoder for 50 epochs with a learning rate of 2e-5, then pick the epoch that performs the best on the validation set for the final usage. Results are shown in Table \ref{tab:two evaluation methods}.

As shown in Table \ref{tab:two evaluation methods}, it is not surprising that fine-tuning leads to the highest clean accuracy score across the board, since it optimizes all the model parameters simultaneously after SCAT pre-training. However, the clean performance gap between fine-tuning and linear evaluation is only 3.7\% on average. In contrast, linear evaluation outperforms fine-tuning under all the robustness-related metrics with large gaps. For example, linear evaluation beats fine-tuning by an average after-attack accuracy of 16.7\%, probably because during fine-tuning, robust parameters in the pre-trained SCAT encoder will be updated to fit the clean examples, while linear evaluation will not modify these robust parameters. Moreover, linear evaluation's computational demand is much smaller than fine-tuning, especially for huge pre-trained language models with billions of parameters. For the above reasons, we choose to adopt linear evaluation for our experiments and we regard how to better combine fine-tuning with our SCAT framework as a future research direction to pursue.

\begin{table}[tb]
\renewcommand\arraystretch{1.1}
  \centering
   \setlength{\tabcolsep}{5mm}{
    \small
    \begin{tabular}{cccc}
    \toprule[1pt]
      \textbf{Dataset} & \textbf{Train} & \textbf{Dev} & \textbf{Test} 
      \\
     \midrule[1pt]
     AG & 120000 & 2000 & 1000   \\
     DBPedia & 560000 & 2000 & 1000 \\
     \bottomrule[1pt]
    \end{tabular}
  }    
    \caption{Statistics of the two text classification datasets we used.}
  \label{tab:dataset}
\end{table}

\section{Descriptions and Statistics of Datasets}
\label{appendix:stats}
In this section, we provide the detailed descriptions and statistics of the two datasets we used in experiments:
\begin{itemize}
\setlength{\itemsep}{0pt}
\setlength{\parsep}{0pt}
\setlength{\parskip}{0pt}
    \item \textbf{AG's News (AG):} Sentence-level classification task with 4 news-type categories: World, Sport, Business, and Science/Technology \cite{zhang2015character}. For each news article, we concatenate its title and description field as the input for our encoders following the setting of \citet{jin2020bert}.
    \item \textbf{DBPedia\footnote{\url{https://www.dbpedia.org/}}:} Extracted from Wikipedia, DBPedia is a sentence-level classification dataset containing 14 non-overlapping categories \cite{zhang2015character}. We also concatenate the title and abstract as the input for each Wikipedia article.
\end{itemize}
\tref{tab:dataset} summarizes the statistics of the two datasets.
 
\section{Detailed Descriptions of two Attackers}
\label{appendix:attacker}
\begin{itemize}
\setlength{\itemsep}{0pt}
\setlength{\parsep}{0pt}
\setlength{\parskip}{0pt}
    \item \textbf{TextFooler\footnote{\url{ https://github.com/jind11/TextFooler}}:} Proposed by \citet{jin2020bert}, given a sentence, TextFooler replaces the tokens with their synonyms extracted from the counter-fitting word embedding \cite{mrkvsic2016counter}. 
    It also uses similarity scores to filter out the low-quality examples to control the consistency between the generated adversarial examples and their original counterparts.
    \item \textbf{BERT-Attack\footnote{\url{https://github.com/LinyangLee/BERT-Attack}}:} Another powerful token replacement-based attack designed by \citet{li2020bert}. For each target token, BERT-Attack first applies a pre-trained BERT to generate all potential replacing candidates, then substitutes the tokens greedily based on the change of model predictions.
\end{itemize}

\section{Details about the Calculation of Average Results in \sref{main}}
\label{appendix:calculation}
In \sref{main}, we mention that ``For both datasets, SCAT-BERT and SCAT-Transformer outperform BERT and Transformer by an average attack failure rate of 24.9\% against TextFooler and 24.1\% against BERT-Attack''. For clarity, we provide details of the calculation here. For TextFooler, looking at rows 1, 3, 4, and 6 of Table \ref{tab:main results}, we have $24.9\% \approx [(28.1-0.3) + (49.0-17.4) + (19.3-8.1) + (49.0-20.2)] / 4$. For BERT-Attack,  looking at rows 7, 9, 10, and 12, we have $24.1\% \approx [(19.8-0.5) + (48.6-3.7) + (34.5-20.8) + (63.7-43.8) + (11.2-0.6) + (49.1-4.7) + (27.7-16.9) + (75.5-46.4)]  / 8$.

In \sref{main}, we also mention that ``As shown in Table \ref{tab:main results}, SCAT enhances Transformer and BERT's performance on after-attack accuracy by an average of 31.7\% against BERT-Attack with antonym filtering''. Looking at rows 7, 9, 10 and 12 of Table~\ref{tab:main results}, we have $31.7\% \approx [(44.0-3.4) + (45.7-4.6) + (58.7-41.7) + (74.3-46.1)] / 4$.

\begin{table}[tb]
\renewcommand\arraystretch{1.1}
  \centering
   \setlength{\tabcolsep}{1.2mm}{
    \small
    \begin{tabular}{cl|*3{c}}
    \toprule[1pt]
      \textbf{Attacker} & \textbf{Model} & \textbf{Acc} & \textbf{Atk Acc} & \textbf{AFR} 
      \\
     \midrule[1pt]
     \multirow{2}*{TextFooler} 
     ~ & Mean & 92.2 & 45.1 & 49.0 
     \\
     ~ & SD & 0.46 & 1.30 & 1.38 
     \\
     \midrule[1pt]
     \multirow{2}*{BERT-Attack} 
     ~ & Mean & 92.2 & 31.7/58.7  & 34.5/63.7 
     \\
     ~ & SD & 0.46 & 1.20/0.85  & 1.39/0.95 
     \\
     \bottomrule[1pt]
    \end{tabular}
  }    
    \caption{Mean and standard deviation (SD) of SCAT-BERT's results among its three different runs on AG. Complete results are shown in Appendix~\ref{appendix:complete main}.}
  \label{tab:mean}%
\end{table}

\begin{table*}[tb]
\renewcommand\arraystretch{1.1}
  \centering
   \setlength{\tabcolsep}{1.5mm}{
    \small
    \begin{tabular}{cl|*3{c}|*3{c}}
    \toprule[1.5pt]
    \multicolumn{2}{c|}{} & \multicolumn{3}{c|}{\textbf{AG}} & \multicolumn{3}{c}{\textbf{DBPedia}} \\
      \textbf{Attacker} & \textbf{Model} & \textbf{Acc} & \textbf{Atk Acc} & \textbf{AFR} 
      & \textbf{Acc} & \textbf{Atk Acc} & \textbf{AFR} 
      \\
     \midrule[1.5pt]
     \multirow{18}*{TextFooler} & Transformer & 92.3 & 0.3 & 0.3 
     & 98.7 & 8.7 & 8.1 
     \\
     \cline{2-8}
     ~ & CL-Transformer(s1) & 84.6 & 13.1 & 15.5 
     & 94.1 & 12.5 & 13.3 
     \\
      ~ & CL-Transformer(s2) & 88.4 & 14.0 & 15.8 
     & 93.6 & 13.0 & 13.9 
     \\
      ~ & CL-Transformer(s3) & 90.1 & 16.5 & 18.3 
     & 93.3 & 13.2 & 14.2 
     \\
     \cline{3-8}
     ~ & SD & 2.30 & 1.44 & 1.26 
     & 0.33 & 0.29 & 0.37 
     \\
     \cline{2-8}
     ~ & SCAT-Transformer(s1) & 90.4 & 26.3 & 29.1 
     & 94.0 & 17.3 & 18.4 
     \\
     ~ & SCAT-Transformer(s2) & 90.1 & 24.4 & 27.1 
     & 90.0 & 17.6 & 19.6 
     \\
     ~ & SCAT-Transformer(s3) & 90.8 & 25.5 & 28.1 
     & 94.6 & 18.9 & 20.0 
     \\
     \cline{3-8}
     ~ & SD & 0.29 & 0.78 & 0.82 
     & 2.04 & 0.69 & 0.68 
     \\
     \cline{2-8}
     ~ & BERT & 95.3 & 16.6 & 17.4 
     & 99.4 & 20.1 & 20.2 
     \\
     \cline{2-8}
     ~ & CL-BERT(s1) & 92.4 & 37.2 & 40.3 
     & 99.0 & 31.6 & 31.9 
     \\
     ~ & CL-BERT(s2) & 92.9 & 40.1 & 43.2 
     & 98.9 & 29.4 & 29.7 
     \\
      ~ & CL-BERT(s3) & 92.2 & 37.0 & 40.1 
     & 98.8 & 36.4 & 36.8 
     \\
     \cline{3-8}
      ~ & SD & 0.29 & 1.42 & 1.42 
     & 0.08 & 2.92 & 2.97 
     \\
      \cline{2-8}
     ~ & SCAT-BERT(s1) & 91.7 & 43.8 & 47.8 
     & 98.3 & 48.5 & 49.3 
     \\
     ~ & SCAT-BERT(s2) & 92.8 & 44.7 & 48.2 
     & 98.7 & 48.0 & 48.6 
     \\
     ~ & SCAT-BERT(s3) & 92.1 & 46.9 & 50.9 
     & 98.5 & 48.4 & 49.1 
     \\
     \cline{3-8}
      ~ & SD & 0.46 & 1.30 & 1.38 
     & 0.16 & 0.22 & 0.29 
     \\
     \midrule[1.5pt]

     \multirow{18}*{BERT-Attack} & Transformer & 92.3 & 0.5/3.4 & 0.5/3.7 
     & 98.7 & 0.6/4.6 & 0.6/4.7 
     \\
     \cline{2-8}
     ~ & CL-Transformer(s1) & 84.6 & 11.9/21.7 & 14.1/25.7 
     & 94.1 & 10.2/26.4 & 10.8/28.0 
     \\
     ~ & CL-Transformer(s2) & 88.4 & 12.9/25.9 & 14.6/29.3 
     & 93.6 & 7.0/19.2 & 7.5/20.5 
     \\
     ~ & CL-Transformer(s3) & 90.1 & 15.3/28.9 & 17.0/32.1 
     & 93.3 & 9.7/24.5 & 10.4/26.3 
     \\
     \cline{3-8}
     ~ & SD & 2.30 & 1.43/2.95 & 1.27/2.62 
     & 0.33 & 1.41/3.05 & 1.47/3.21 
     \\
     \cline{2-8}
     ~ & SCAT-Transformer(s1) & 90.4 & 19.1/44.6 & 21.1/49.3 
     & 94.0 & 9.3/49.4 & 9.9/52.6 
     \\
     ~ & SCAT-Transformer(s2) & 90.1 & 17.8/43.9 & 19.8/48.7 
     & 90.0 & 10.8/36.4 & 12.0/40.4 
     \\
     ~ & SCAT-Transformer(s3) & 90.8 & 16.8/43.5 & 18.5/47.9 
     & 94.6 & 11.0/51.4 & 11.6/54.3 
     \\
     \cline{3-8}
     ~ & SD & 0.29 & 0.94/0.46 & 1.06/0.57 
     & 2.04 & 0.76/6.65 & 0.91/6.19 
     \\
     \cline{2-8}
     ~ & BERT & 95.3 & 19.8/41.7 & 20.8/43.8 
     & 99.4 & 16.8/46.1 & 16.9/46.4 
     \\
     \cline{2-8}
     ~ & CL-BERT(s1) & 92.4 & 33.6/50.3 & 36.4/54.4 
     & 99.0 & 24.8/56.7 & 25.1/57.3 
     \\
     ~ & CL-BERT(s2) & 92.9 & 38.7/54.1 & 41.7/58.2 
     & 98.9 & 21.1/53.8 & 21.3/54.4 
     \\
     ~ & CL-BERT(s3) & 92.2 & 38.1/52.3 & 41.3/56.7 
     & 98.8 & 27.6/58.5 & 27.9/59.2 
     \\
     \cline{3-8}
     ~ & SD & 0.29 & 2.28/1.55 & 2.41/1.56 
     & 0.08 & 2.66/1.94 & 2.71/1.97 
     \\
      \cline{2-8}
     ~ & SCAT-BERT(s1) & 91.7 & 31.5/57.9 & 34.4/63.1 
     & 98.3 & 27.1/75.1 & 27.6/76.4 
     \\
     ~ & SCAT-BERT(s2) & 92.8 & 30.4/58.4 & 32.8/62.9 
     & 98.7 & 25.5/74.1 & 25.8/75.1 
     \\
     ~ & SCAT-BERT(s3) & 92.1 & 33.3/59.9 & 36.2/65.0 
     & 98.5 & 29.1/73.8 & 29.5/74.9 
     \\
     \cline{3-8}
     ~ & SD & 0.46 & 1.20/0.85 & 1.39/0.95 
     & 0.16 & 1.47/0.56 & 1.51/0.67 
     \\
     \bottomrule[1.5pt]
    \end{tabular}
  }    
    \caption{Complete results of our main experiment (Table \ref{tab:main results}). We also show the standard deviation (SD) values of each model's performance while the average values are listed in Table \ref{tab:main results}. Put inside brackets, s1, s2 and s3 denote the three different seeds used for three runs.}
  \label{tab:complete main results}%
\end{table*}

\begin{table*}[tb]
\renewcommand\arraystretch{1.1}
  \centering
   \setlength{\tabcolsep}{3mm}{
    \small
    \begin{tabular}{lc|*2{c}|*2{c}}
    \toprule[1.5pt]
     & & \multicolumn{2}{c|}{\textbf{TextFooler}} & \multicolumn{2}{c}{\textbf{BERT-Attack}} \\
       \textbf{Model} & \textbf{Acc} & \textbf{Atk Acc} & \textbf{AFR} 
       & \textbf{Atk Acc} & \textbf{AFR} 
      \\
     \midrule[1.5pt]
      Transformer & 92.3  & 0.3 & 0.3 
      & 0.5/3.4 & 0.5/3.7 
      \\
      \cline{1-6}
      +CL(s1) &84.6 & 13.1 & 15.5 & 11.9/21.7 & 14.1/25.7  \\
      +CL(s2) & 88.4 &14.0  &15.8  &12.9/25.9  &14.6/29.3  \\
      +CL(s3) & 90.1 &16.5  &18.3  &15.3/28.9  &17.0/32.1  \\
      \cline{2-6}
      SD & 2.30 & 1.44 & 1.26 & 1.43/2.95 & 1.27/2.62 \\
       \cline{1-6}
     +CL+Extra Aug(s1) & 89.5 & 19.1 & 21.3 &18.2/31.0 &20.3/34.6 \\
     +CL+Extra Aug(s2) &89.0  &20.9  &23.5  &16.8/31.8 &18.9/35.7 \\
     +CL+Extra Aug(s3) &87.6  &19.1  &21.8  &18.0/30.7 &20.6/35.1 \\
     \cline{2-6}
     SD & 0.80 &0.85  &0.94  &0.62/0.46 &0.74/0.45 \\
      \cline{1-6}
     +SCAT(Random)(s1) &91.5 &21.5 &23.5 &18.3/43.5 &20.0/47.5 \\
     +SCAT(Random)(s2) &91.7 &22.1 &24.1 &20.8/44.9 &22.7/49.0 \\
     +SCAT(Random)(s3) &91.3 &22.3 &24.4 &14.8/39.6 &16.2/43.4 \\
     \cline{2-6}
     SD &0.16 &0.34 &0.37 &2.46/2.24 &2.67/2.37 \\
      \cline{1-6}
     +SCAT(s1) &90.4 &26.3 &29.1 &19.1/44.6 &21.1/49.3 \\
     +SCAT(s2) &90.1 &24.4 &27.1 &17.8/43.9 &19.8/48.7 \\
     +SCAT(s3) &90.8 &25.5 &28.1 &16.8/43.5 &18.5/47.9 \\
     \cline{2-6}
     SD &0.29 &0.78 &0.82 &0.94/0.46 &1.06/0.57 \\
     \midrule[1.5pt]

     BERT & 95.3  & 16.6 & 17.4 
      & 19.8/41.7 & 20.8/43.8 
      \\
       \cline{1-6}
      +CL(s1) & 92.4 & 37.2 & 40.3 & 33.6/50.3 & 36.4/54.4 \\
      +CL(s2) & 92.9 & 40.1 & 43.2 & 38.7/54.1 & 41.7/58.2 \\
      +CL(s3) & 92.2 & 37.0 & 40.1 & 38.1/52.3 & 41.3/56.7 \\
      \cline{2-6}
      SD &0.29  &1.42  &1.42  &2.28/1.55  &2.41/1.56  \\
       \cline{1-6}
     +CL+Extra Aug(s1) & 92.7 & 42.0 & 45.3 &40.7/53.3 &43.9/57.5 \\
     +CL+Extra Aug(s2) & 92.9 & 46.7 & 50.3 &43.8/57.8 &47.2/62.2 \\
     +CL+Extra Aug(s3) & 92.4 &44.6  &48.3  &41.1/56.5 &44.5/61.2 \\
     \cline{2-6}
     SD &0.21  &1.92  &2.06  &1.38/1.89 &1.44/2.02 \\
      \cline{1-6}
     +SCAT(Random)(s1) &91.9 &39.0 &42.4 &37.0/61.1 &40.3/66.5 \\
     +SCAT(Random)(s2) &92.2 &41.8 &45.3 & 35.3/61.2&38.3/66.4 \\
     +SCAT(Random)(s3) &92.3 &43.6 &47.2 &40.2/63.4 &43.6/68.7 \\
     \cline{2-6}
     SD &0.17 &1.89 &1.97 &2.03/1.06 &2.19/1.06 \\
      \cline{1-6}
     +SCAT(s1) &91.7 &43.8 &47.8 &31.5/57.9 &34.4/63.1 \\
     +SCAT(s2) &92.8 &44.7 &48.2 &30.4/58.4 &32.8/62.9 \\
     +SCAT(s3) &92.1 &46.9 &50.9 &33.3/59.9 &36.2/65.0 \\
     \cline{2-6}
     SD &0.46 &1.30 &1.38 &1.20/0.85 &1.39/0.95 \\
     \bottomrule[1.5pt]
    \end{tabular}
  }
    \caption{
    Complete results of the ablation study. We use ``+" here to distinguish different models for brevity. We also show the standard deviation (SD) values of each model's performance while the average values are listed in the main paper. Put inside brackets, s1, s2 and s3 denote the three different seeds used for three runs.}
  \label{tab:complete ablation}
\end{table*}

\section{Complete Results of the Main Experiment}\label{appendix:complete main}

While we only report the average results for the main experiment in Table \ref{tab:main results}, we further report the complete results here for a more complete view of the stability of different models. \tref{tab:complete main results} summarizes all the results.

\section{Complete Results of the Ablation Study}
\label{appendix:complete ablation}
Similar to what we have done for the main experiment, we also report the complete results of the ablation study here for reference. \tref{tab:complete ablation} summarizes all the results.

\section{Mean and Sstandard Ddeviation (SD) of SCAT-BERT's Rresults}
\label{appendix:mean and sd}
In this section, we list the mean and standard deviation of SCAT-BERT's results among its three runs on AG in Table \ref{tab:mean} as a supplement to \sref{sup-adv}.

\end{document}